\documentclass[sigconf]{acmart}
\usepackage{algorithm}
\usepackage{algpseudocode}
\usepackage{float}
\usepackage{multirow}
\usepackage{booktabs}
\usepackage{colortbl}
\usepackage[table]{xcolor}

\definecolor{lightgreen}{RGB}{220,245,220}
\definecolor{lightgray}{RGB}{230,230,230}

\newcommand{\best}[1]{\textbf{#1}}

\newcommand{\consis}{c}   
\newcommand{\anom}{a}     

\copyrightyear{2026}
\acmYear{2026}
\setcopyright{cc}
\setcctype{by-nc-nd}
\acmConference[WWW '26]{Proceedings of the ACM Web Conference 2026}{April 13--17, 2026}{Dubai, United Arab Emirates}
\acmBooktitle{Proceedings of the ACM Web Conference 2026 (WWW '26), April 13--17, 2026, Dubai, United Arab Emirates}
\acmPrice{}
\acmDOI{10.1145/3774904.3792482}
\acmISBN{979-8-4007-2307-0/2026/04}

\title{AC\textsuperscript{2}L-GAD: Active Counterfactual Contrastive Learning for Graph Anomaly Detection}

\author{Kamal Berahmand}
\email{kamal.berahmand@rmit.edu.au}
\orcid{0000-0003-4459-0703}
\affiliation{%
  \institution{RMIT University}
  \city{Melbourne}
  \state{Victoria}
  \country{Australia}
}

\author{Saman Forouzandeh}
\email{saman.forouzandeh@rmit.edu.au}
\orcid{0000-0002-5952-156X}
\affiliation{%
  \institution{RMIT University}
  \city{Melbourne}
  \state{Victoria}
  \country{Australia}
}

\author{Mehrnoush Mohammadi}
\email{m.mohammadibolbanabad@uq.edu.au}
\orcid{0000-0001-9596-7414}
\affiliation{%
  \institution{The University of Queensland}
  \city{Brisbane}
  \state{Queensland}
  \country{Australia}
}

\author{Parham Moradi}
\email{parham.moradi@rmit.edu.au}
\orcid{0000-0002-5604-565X}
\affiliation{%
  \institution{RMIT University}
  \city{Melbourne}
  \state{Victoria}
  \country{Australia}
}

\author{Mahdi Jalili}
\email{mahdi.jalili@rmit.edu.au}
\orcid{0000-0002-0517-9420}
\affiliation{%
  \institution{RMIT University}
  \city{Melbourne}
  \state{Victoria}
  \country{Australia}
}

\begin{document}

\begin{abstract}
Graph anomaly detection aims to identify abnormal patterns in networks, but faces significant challenges from label scarcity and extreme class imbalance. While graph contrastive learning offers a promising unsupervised solution, existing methods suffer from two critical limitations: random augmentations break semantic consistency in positive pairs, while naive negative sampling produces trivial, uninformative contrasts. We propose AC$^2$L-GAD, an Active Counterfactual Contrastive Learning framework that addresses both limitations through principled counterfactual reasoning. By combining information-theoretic active selection with counterfactual generation, our approach identifies structurally complex nodes and generates anomaly-preserving positive augmentations alongside normal negative counterparts that provide hard contrasts, while restricting expensive counterfactual generation to a strategically selected subset. This design reduces computational overhead by approximately 65\% compared to full-graph counterfactual generation while maintaining detection quality. Experiments on nine benchmark datasets, including real-world financial transaction graphs from GADBench, show that AC$^2$L-GAD achieves competitive or superior performance compared to state-of-the-art baselines, with notable gains in datasets where anomalies exhibit complex attribute-structure interactions.

\noindent\textbf{Code:} \url{https://anonymous.4open.science/r/AC2L-GAD-33B8}
\end{abstract}

\keywords{Graph anomaly detection, Contrastive learning, Counterfactual reasoning, Active learning}

\maketitle

\section{Introduction}

Graph anomaly detection (GAD) is a fundamental task in network analysis, with applications in financial fraud prevention, cybersecurity, and biological network analysis. Following prior studies~\cite{zheng2025cluster,li2024diffgad}, we focus on node-level anomaly detection. This task faces two key challenges: (i) the scarcity of labeled anomaly data, as obtaining reliable human annotations is costly and time-consuming, and (ii) severe class imbalance, where anomalies represent only a small fraction of instances while misclassification costs vary asymmetrically across applications. These constraints limit the applicability of supervised methods and have motivated growing interest in unsupervised approaches that leverage abundant unlabeled data~\cite{zheng2025cluster,qiao2025deep}.

Within unsupervised GAD, graph contrastive learning (GCL) has emerged as the dominant paradigm. Unlike reconstruction-based approaches that suffer from ambiguous optimization objectives, unified latent spaces inadequate for diverse anomaly patterns, and high sensitivity to sparse attributes~\cite{ding2019dominant,fan2020anomalydae,peng2020alarm}, GCL frameworks learn discriminative node embeddings by maximizing agreement between semantically consistent views while separating dissimilar views. This paradigm has proven effective across multiple domains~\cite{you2020graph,zhu2020deepgcl}. Recent specialized architectures have built on this foundation: CoLA~\cite{liu2021cola} introduced node-subgraph contrast for local structural coherence, ANEMONE~\cite{jin2021anemone} incorporated multi-scale contrasts, and AD-GCL~\cite{xu2025revisiting} explored anomaly-aware view construction and hard negative mining.

Despite these advances, existing GCL methods face two critical limitations. First, current approaches use stochastic augmentations---including random edge dropping, node removal, and feature masking---to construct positive pairs~\cite{you2020graph,xia2022simgrace,ding2022data}. While effective for general representation learning, such augmentations lack semantic awareness: they apply uniform perturbations without considering local graph structure or anomaly-relevant patterns. Consequently, informative yet rare edges may be removed while spurious connections are retained, degrading semantic consistency and blurring boundaries between normal and abnormal nodes~\cite{hu2025higher,tang2024perturbation}. We term this Gap G1: inconsistent positives due to random augmentation.

Second, negative sampling in existing methods typically relies on distance-based heuristics or uniform random selection~\cite{duan2023graph,ding2022data}. Such strategies produce trivial negatives that are easily separable from anchors, providing weak supervision and yielding coarse decision boundaries. This is particularly problematic in anomaly detection, where anomalies often exhibit subtle deviations, hide within normal patterns, or concentrate in sparse regions~\cite{yang2023generating,liao2024revgnn}. Without exposure to challenging negatives, models fail to develop the discriminative capacity required for accurate detection. We term this Gap G2: uninformative negatives from naive sampling.

These two gaps fundamentally limit the effectiveness of current GCL-based anomaly detection methods. Counterfactual reasoning offers a principled solution to both. Through minimal and controlled perturbations, counterfactual samples preserve semantic validity while introducing targeted variations. For positive pairs, counterfactuals preserve anomalous characteristics through controlled augmentations that maintain structural irregularity and attribute deviation, while for negatives, they normalize nodes by aligning features toward neighborhood centroids and increasing homophily with similar neighbors~\cite{yang2023generating}, thereby addressing inconsistent positives (Gap G1) and uninformative negatives (Gap G2). However, counterfactual generation is computationally expensive, typically requiring iterative gradient-based optimization for each node or subgraph~\cite{lucic2022cfgnnexplainer,long2025adversarial}. Applying this procedure graph-wide results in $O(|V|)$ complexity, making it impractical for large-scale networks.

To overcome this computational bottleneck, we introduce active counterfactual learning, where counterfactuals are generated only for a strategically selected subset of highly informative nodes. Drawing on principles from active learning~\cite{cai2017active,gao2018active}, we design a selection criterion combining structural complexity and attribute divergence to identify nodes that contribute most to representation quality. This reduces computational overhead from $O(|V|)$ to $O(k)$ where $k \ll |V|$, while maintaining---and often improving---model performance by focusing counterfactual generation on the most important training instances. This represents the first integration of active learning with counterfactual reasoning for graph anomaly detection.

We present AC$^2$L-GAD (Active Counterfactual Contrastive Learning for Graph Anomaly Detection), a unified framework that addresses Gaps G1 and G2 through the combination of information-theoretic active selection and principled counterfactual generation. The framework integrates active node selection, counterfactual augmentation, graph encoding, contrastive optimization, and neighborhood-based anomaly scoring into a coherent pipeline, enabling both efficient training and accurate detection. Our contributions are threefold:

\begin{itemize}
    \item We propose AC$^2$L-GAD, a unified framework that integrates active learning with counterfactual reasoning to address two fundamental limitations in graph contrastive learning for anomaly detection: Gap G1 (inconsistent positives from random augmentation) and Gap G2 (uninformative negatives from naive sampling), achieving both representational quality and computational efficiency.
    \item We introduce an active counterfactual generation mechanism that combines information-theoretic node selection with principled counterfactual reasoning to 
generate anomaly-preserving positive augmentations and normalized hard negatives, reducing computational overhead by approximately 65\% compared to full-graph 
counterfactual generation while maintaining detection quality.
    \item We conduct extensive experiments across nine benchmark datasets, including real-world financial fraud graphs from GADBench, demonstrating that AC$^2$L-GAD achieves competitive or superior performance compared to eighteen state-of-the-art baselines, with notable improvements on datasets where anomalies exhibit complex attribute-structure interactions.
\end{itemize}


\section{Related Work}

Graph anomaly detection addresses the challenge of distinguishing abnormal nodes hidden among normal ones. Existing methods fall into three paradigms: traditional statistical approaches, reconstruction-based methods, and contrastive learning frameworks.

\subsection{Traditional Graph Anomaly Detection}

Traditional approaches rely on hand-crafted features and shallow heuristics. 
LOF~\cite{breunig2000lof} employs density-based estimation, comparing a node’s local reachability density to that of its neighbors. AMEN~\cite{perozzi2016amen} ranks anomalous attributed neighborhoods by jointly assessing structural consistency and attribute agreement in egonets. RADAR~\cite{li2017radar} formulates anomaly scoring as a regularized residual analysis over structure–attribute inconsistencies, while ANOMALOUS~\cite{peng2018anomalous} jointly models structure and attributes to uncover deviations. 
GUIDE~\cite{yuan2021guide} leverages higher-order structural motifs (motif-degree) with a dual-autoencoder to reconstruct attributes and higher-order structures for unsupervised detection. While efficient and interpretable, these methods are limited by linear assumptions and manual feature engineering, and struggle with complex dependencies in high-dimensional attributed graphs, trading scalability and generalization for interpretability.

\subsection{Reconstruction-based Methods}

Reconstruction-based methods detect anomalies through reconstruction discrepancies, assuming that anomalous nodes are harder to reconstruct accurately due to their deviation from normal structural and attribute patterns. 
DOMINANT~\cite{ding2019dominant} first proposed a GCN-based autoencoder framework to jointly reconstruct attributes and topology. 
AnomalyDAE~\cite{fan2020anomalydae} utilizes a dual denoising autoencoder to enhance robustness against noisy features, while 
ALARM~\cite{peng2020alarm} adopts local reconstruction to capture fine-grained structural and attribute inconsistencies. 
ComGA~\cite{luo2022comga} introduces community-aware reconstruction for topology-attribute alignment. 
ADA-GAD~\cite{he2024adag} employs an adaptive denoising mechanism that dynamically balances structural and attribute information. 
Finally, GAD-NR~\cite{roy2024gadnr} performs neighborhood-level reconstruction using residual learning to improve sensitivity to local irregularities. 
Although these models effectively capture reconstruction errors, they remain sensitive to noise and often fail to generalize across diverse anomaly types.

\subsection{Graph Contrastive Learning Methods}

Graph contrastive learning has emerged as a powerful paradigm for unsupervised graph anomaly detection (GAD), aiming to learn representations that make normal node-context pairs similar and abnormal ones dissimilar. CoLA~\cite{liu2021cola} pioneered node-subgraph contrast, where a node and its RWR-sampled local subgraph form a positive pair. ANEMONE~\cite{jin2021anemone} extended this with a node-node branch to capture multi-scale structural and contextual irregularities. Subsequent works refined contrasting views and supervision: Sub-CR~\cite{zhang2022subcr} decoupled local and global structures through intra-/inter-view contrast enhanced by attribute-consistency reconstruction, while GRADATE~\cite{duan2023gradate} introduced subgraph-subgraph contrast to strengthen representation robustness. More recently, AD-GCL~\cite{xu2025revisiting} re-examined GCL-based anomaly detection from a structural imbalance perspective, proposing neighbor pruning for head nodes and anomaly-guided neighbor completion for tail nodes. In distributed settings, FedCAD~\cite{kong2025fedcad} pioneered federated contrastive GAD via pseudo-labeled anomaly exchange and cross-client aggregation, while FedCLGN~\cite{wu2025fedclgn} improved privacy-preserving detection through global negative-pair pooling and local graph diffusion.


\section{Preliminaries}

\subsection{Attributed Graph and Notation}
We consider an \emph{attributed graph} defined as $G=(V,E,X)$, where $V$ is the set of $n=|V|$ nodes, $E$ is the set of edges, and $X \in \mathbb{R}^{n \times d}$ is the node feature matrix with $d$-dimensional attributes for each node. The encoder $f_\theta$ maps nodes to low-dimensional embeddings that preserve both structural and attribute information.  
Table~\ref{tab:notation} summarizes the main symbols used throughout the paper.

\begin{table}[h]
\centering
\caption{Summary of notations.}
\small
\label{tab:notation}
\begin{tabular}{ll}
\toprule
\textbf{Symbol} & \textbf{Description} \\
\midrule
$G=(V,E,X)$ & Attributed graph (nodes, edges, features) \\
$n=|V|$ & Number of nodes \\
$d$ & Feature dimension \\
$X \in \mathbb{R}^{n \times d}$ & Node feature matrix \\
$f_\theta(\cdot)$ & Graph encoder parameterized by $\theta$ \\
$z_i$ & Embedding of node $v_i$ \\
$S \subset V$ & Selected subset ($|S|=k$) \\
$\mathcal{H}_{\text{topo}}(v_i)$ & Topology entropy \\
$\mathcal{D}(v_i)$ & Attribute deviation \\
$\consis(v_i)$ & Neighborhood consistency score \\
$\anom(v_i)$ & Anomaly score in embedding space \\
$x_i^+, x_i^-$ & Positive/negative \emph{counterfactual} features \\
$E_i^+, E_i^-$ & Positive/negative \emph{counterfactual} edges \\
$\lambda_u$ & Uniformity regularization weight \\
\bottomrule
\end{tabular}
\end{table}

\subsection{Unsupervised Graph Anomaly Detection}
The objective of unsupervised GAD is to learn a scoring function
\[
f: V \times G \;\rightarrow\; \mathbb{R},
\]
that assigns each node $v_i \in V$ an anomaly score $f(v_i,G)$, where larger scores correspond to higher likelihoods of being anomalous. Unlike supervised methods, no anomaly labels are available during training, making the task highly challenging due to the scarcity of labeled anomalies and the extreme imbalance between normal and abnormal instances. Existing approaches either rely on reconstruction errors (e.g., autoencoders) or embedding deviations (e.g., consistency with local neighborhoods), both of which motivate our counterfactual-based formulation.

\subsection{Graph Contrastive Learning}
Graph Contrastive Learning (GCL) has emerged as a powerful paradigm for unsupervised representation learning on graphs ~\cite{you2020graph}. It constructs augmented views of the same node or subgraph and maximizes their agreement while contrasting against negatives.  

Given two augmentation distributions $\mathcal{T}_1$ and $\mathcal{T}_2$, with $t_1(G)\sim\mathcal{T}_1(G)$ and $t_2(G)\sim\mathcal{T}_2(G)$, the standard InfoNCE objective is:
\begin{equation}
\mathcal{L}_{\text{GCL}} = -\sum_{i \in V} \log 
\frac{\exp(\mathrm{sim}(z_i^{(1)}, z_i^{(2)})/\tau)}
{\sum_{j \in V} \exp(\mathrm{sim}(z_i^{(1)}, z_j^{(2)})/\tau)},
\end{equation}
where $z_i^{(m)} = f_\theta(t_m(G), v_i)$ is the embedding of node $i$ under view $m$, and $\mathrm{sim}(\cdot,\cdot)$ denotes the normalized dot product (cosine similarity). In practice, negatives are drawn from in-batch samples for scalability, rather than the entire graph.

\subsection{Counterfactual Reasoning in Graphs}

Counterfactual reasoning provides a principled framework for generating informative training samples through minimal, controlled perturbations~\cite{lucic2022cfgnnexplainer,vu2022counterfactual}. Given a node $v_i$ with feature vector $x_i$ and neighborhood $N(i)$, we define counterfactuals as minimal perturbations that alter the node's alignment with its local context, measured through a consistency score $c(v_i)$ (detailed in Section 4.2):
\begin{align}
x_i^{+} &= \arg\min_{\Delta x} \|\Delta x\|_2 \quad \text{s.t.} \quad c(v_i + \Delta x) > \gamma \cdot c(v_i), \quad \gamma > 1, \\
x_i^{-} &= \arg\min_{\Delta x} \|\Delta x\|_2 \quad \text{s.t.} \quad c(v_i + \Delta x) < \beta \cdot c(v_i), \quad \beta \in (0,1).
\end{align}

Positive counterfactuals $x_i^{+}$ preserve anomalous characteristics by amplifying inconsistency with the neighborhood, while negative counterfactuals $x_i^{-}$ normalize the node by reducing deviation toward its local context. This generates semantically similar positives and dissimilar hard negatives for effective contrastive learning.

\section{Method: AC$^2$L-GAD}

We introduce AC$^2$L-GAD (Active Counterfactual Contrastive Learning for Graph Anomaly Detection), an unsupervised framework that addresses two fundamental limitations in graph contrastive learning: (G1) random positive sampling that disrupts semantic consistency, and (G2) trivial negative sampling that provides weak supervision.

\begin{figure*}[!t]
  \centering
  \includegraphics[width=\textwidth,keepaspectratio]{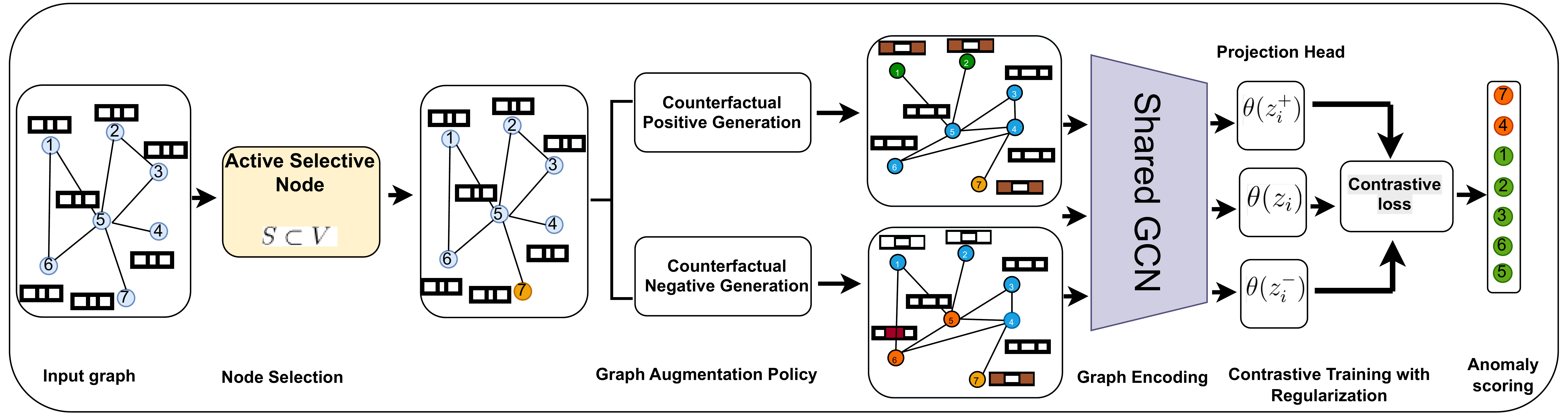}
  \caption{AC$^2$L-GAD pipeline. 
  The framework performs active node selection to focus on informative regions, generates anomaly-preserving counterfactual positives and normalized counterfactual negatives, and encodes original and augmented views with a shared GCN. 
  A contrastive objective with uniformity regularization shapes the embedding space, from which anomaly scores are derived.}
  \label{fig:ac2lgad-pipeline}
\end{figure*}

The framework integrates five modules into a unified pipeline (Fig.~\ref{fig:ac2lgad-pipeline}). Active node selection identifies structurally complex and attribute-divergent nodes to focus computational effort on anomaly-prone regions. Counterfactual augmentation generates positives that preserve anomalous characteristics through controlled variations and negatives that normalize nodes toward local patterns, creating effective anomaly-aware contrasts. A shared GCN encoder with lightweight projection heads processes original and counterfactual views. Contrastive optimization with uniformity regularization aligns anomalous anchors with their anomaly-preserving positives while separating them from normalized negatives. Finally, neighborhood-based scoring quantifies each node's deviation from its local context to produce anomaly scores.

\subsection{Active Node Selection via Structural and Attribute Criteria}

As a preparatory step for \emph{adaptive} augmentation, AC$^2$L-GAD restricts augmentation to a \emph{subset} of nodes rather than the entire graph. Concretely, we select a subset $S \subset V$ with $|S|=k$ and $k \ll |V|$, so that training focuses on anomaly-relevant regions and the per-epoch augmentation/sampling cost drops from $O(|V|)$ to $O(k)$. We set $k=\max(100,\lfloor 0.1|V| \rfloor)$ to balance efficiency and coverage. This design follows the \emph{active learning} principle that most informative signal typically concentrates in a small number of \emph{hard and diverse} samples; hence, operating on $S$ provides a stronger supervision signal for downstream contrastive learning at a fraction of the cost.

\paragraph{Structural Complexity via Local Topology Entropy.}
We introduce a novel measure of structural informativeness that quantifies the heterogeneity of local patterns in a node's neighborhood. For node $v_i$, we discretize structural indicators of its neighbors (degree, clustering coefficient, triangle count) into $B=5$ bins based on graph-wide quantiles, and define:
\begin{equation}
\mathcal{H}_{\text{topo}}(v_i) = -\sum_{b \in B} p(b \,|\, v_i)\,\log p(b \,|\, v_i),
\end{equation}
where $p(b \,|\, v_i)$ is the normalized frequency of pattern $b$ among neighbors of $v_i$. A high entropy value indicates a structurally complex and informative local topology.

\paragraph{Attribute Deviation.}
To complement structural information, we also consider nodes whose attributes strongly deviate from their neighborhood. For node $v_i$ with feature vector $\mathbf{x}_i$ (where all features in $X$ are z-score normalized), we compute:
\begin{equation}
\mathcal{D}(v_i) = \frac{\|\mathbf{x}_i - \bar{\mathbf{x}}_{N(i)}\|_2}{\sigma(\{\mathbf{x}_j:j\in N(i)\}) + \epsilon},
\end{equation}
where $\bar{\mathbf{x}}_{N(i)}$ and $\sigma(\cdot)$ denote the mean and standard deviation of neighbor features, respectively, and $\epsilon = 10^{-6}$ is a small constant for numerical stability. Large values signal attribute-level anomalies at decision boundaries.

\subsubsection{Final Selection Rule}
We select the top-$k/2$ nodes ranked by $\mathcal{H}_{\text{topo}}$ and the top-$k/2$ nodes ranked by $\mathcal{D}$, and take their union as $S$, resulting in $|S| \leq k$ nodes (with $|S| = k$ when the sets are disjoint). This procedure runs in $O(|V|d\bar{d})$ time and is executed once before training. The dual criterion is robust: entropy favors structurally heterogeneous regions, while deviation highlights attribute-level boundaries. In our ablation study we further validate the individual and union-based contributions of the two criteria.


\subsection{Active Counterfactual Generation}

Counterfactual augmentation is applied only to the informative subset $S$ identified in the previous step. Unlike conventional methods that randomly sample negatives, our approach synthesizes counterfactuals that maintain anomalous patterns for positive pairs and normalize nodes for negative pairs, eliminating dependence on ground-truth labels while providing hard, informative contrasts. This directly tackles random positives (G1) and trivial negatives (G2).

Following the counterfactual framework introduced in Section 3.4, we generate positive and negative counterfactuals for each node in $S$ through constrained optimization.

\subsubsection{Feature Counterfactuals}

We formalize counterfactual generation as constrained optimization problems that balance minimal perturbation with sufficient change in neighborhood consistency. To measure consistency, we define a combined score integrating attribute deviation and structural alignment:
\begin{equation}
c(v_i) = \lambda_{\text{attr}} \cdot \frac{\|x_i - \bar{x}_{N(i)}\|_2}{\sigma_{N(i)} + \delta} + \lambda_{\text{struct}} \cdot \left(1 - \frac{|N(i)\cap N_{\text{sim}}(i)|}{|N(i)|}\right),
\end{equation}
where $\bar{x}_{N(i)} = \frac{1}{|N(i)|}\sum_{j \in N(i)} x_j$ is the neighbor feature centroid, $\sigma_{N(i)}$ denotes neighborhood standard deviation, $N_{\text{sim}}(i) = \{j \in N(i) : \cos(x_i, x_j) > 0.7\}$ identifies feature-similar neighbors (threshold chosen empirically), $\lambda_{\text{attr}} + \lambda_{\text{struct}} = 1$, and $\delta = 10^{-6}$ ensures numerical stability. Higher scores indicate greater inconsistency. We set $\lambda_{\text{attr}} = 0.8$ and $\lambda_{\text{struct}} = 0.2$, prioritizing attribute-based consistency as it proved more informative in preliminary validation.

\textbf{Positive Counterfactuals.}
The ideal positive counterfactual solves:
\begin{equation}
x_i^+ = \arg\min_{x'} \|x_i - x'\|_2 \quad \text{s.t.} \quad c(v_i') \ge \gamma \cdot c(v_i), \quad \gamma = 1.3.
\end{equation}
Since exact optimization is intractable due to the non-convex nature of $c(\cdot)$, we employ a gradient-based approximation. Since the first term of $c(v_i)$ (attribute deviation) dominates with $\lambda_{\text{attr}}=0.8$, moving away from the neighbor centroid effectively increases $c(v_i)$:
\begin{equation}
x_i^+ = x_i - \alpha_{\text{pos}} \cdot \frac{\bar{x}_{N(i)} - x_i}{\|\bar{x}_{N(i)} - x_i\|_2+\delta},
\end{equation}
where $\alpha_{\text{pos}} = \min(0.3, (\gamma-1) \cdot \frac{\|x_i - \bar{x}_{N(i)}\|_2}{0.5 \cdot \text{std}(X)})$ is an adaptive step size proportional to the current distance from the centroid, with maximum bound 0.3 to prevent excessive perturbations. We verify $c(v_i^+) > c(v_i)$ and $\|x_i - x_i^+\|_2 \le 0.5 \cdot \text{std}(X)$; if violated, we scale $\alpha_{\text{pos}}$ by 0.5 iteratively (maximum 5 iterations). If constraints cannot be satisfied, we retain the original $x_i$ as a fallback positive (maintaining the anchor's anomalous features without perturbation).

\textbf{Negative Counterfactuals.}
Similarly, negative counterfactuals are generated by:
\begin{equation}
x_i^- = \arg\min_{x'} \|x_i - x'\|_2 \quad \text{s.t.} \quad c(v_i') \le \beta \cdot c(v_i), \quad \beta = 0.7,
\end{equation}
with gradient-based approximation moving toward the neighbor centroid:
\begin{equation}
x_i^- = x_i + \alpha_{\text{neg}} \cdot \frac{\bar{x}_{N(i)} - x_i}{\|\bar{x}_{N(i)} - x_i\|_2+\delta},
\end{equation}
where $\alpha_{\text{neg}} = \min(0.3, (1-\beta) \cdot \frac{\|x_i - \bar{x}_{N(i)}\|_2}{0.5 \cdot \text{std}(X)})$ follows the same adaptive principle. We verify $c(v_i^-) < c(v_i)$ and perturbation bounds, applying iterative scaling if needed (maximum 5 iterations). If constraints cannot be satisfied, we discard the negative sample and rely on in-batch negatives during contrastive training. Empirical validation of approximation quality, constraint satisfaction rates, and failure mode analysis are provided in Appendix B.1.

\subsubsection{Structural Counterfactuals}

In addition to feature modifications, we adjust the local graph structure to enhance counterfactual quality. Positive structural counterfactuals aim to decrease homophily by connecting the node to dissimilar neighbors or removing connections to similar ones:
\begin{equation}
E_i^+ = \arg\min_{E'} |E_i \triangle E'| \quad \text{s.t.} \quad \text{homophily}(v_i,E') < \text{homophily}(v_i,E_i),
\end{equation}
where $E_i$ denotes the set of edges incident to node $v_i$, $E_i \triangle E'$ denotes their symmetric difference, and
\begin{equation}
\text{homophily}(v_i, E) = \frac{|\{j \in N_E(i) : \cos(x_i, x_j) > 0.7\}|}{|N_E(i)|}
\end{equation}
measures the fraction of feature-similar neighbors under edge set $E$. Negative structural counterfactuals increase homophily:
\begin{equation}
E_i^- = \arg\min_{E'} |E_i \triangle E'| \quad \text{s.t.} \quad \text{homophily}(v_i,E') > \text{homophily}(v_i,E_i).
\end{equation}

Since exact optimization is NP-hard, we employ a greedy heuristic: for positive counterfactuals, we iteratively remove up to 2 edges to feature-similar neighbors while optionally adding 1 edge to dissimilar 2-hop neighbors, ensuring degree change $\le 2$ and decreased homophily. For negative counterfactuals, we add up to 2 edges to similar 2-hop neighbors and remove 1 dissimilar edge. The algorithm terminates when constraints are satisfied or no valid modifications remain. Detailed pseudocode and approximation analysis are in Appendix B.2.

The combined counterfactual for node $v_i$ integrates both feature and structural modifications: $v_i^+ = (x_i^+, E_i^+)$ and $v_i^- = (x_i^-, E_i^-)$. During GCN encoding, we create node-specific views where each $v_i \in S$ uses feature vector $x_i^{\pm}$ and a locally modified adjacency matrix in which edges incident to $v_i$ are updated according to $E_i^{\pm}$ while all other edges remain unchanged. Combined quality metrics demonstrating the complementary strengths of feature and structural counterfactuals are presented in Appendix B.3.


\subsection{Graph Encoding and Projection Head}

We encode original and counterfactual views using a shared 2-layer GCN encoder:
\begin{align}
H^{(1)} &= \text{ReLU}(\hat{A}XW^{(0)}), \\
Z &= \hat{A}H^{(1)}W^{(1)},
\end{align}
where $W^{(0)}, W^{(1)}$ are trainable weight matrices with dimensions $d \times 64$ and $64 \times 32$ respectively. For each node $v_i \in S$, we generate three embeddings using view-specific adjacency matrices: $z_i$ (original graph $A$), $z_i^+$ (modified graph with edges $E_i^+$), and $z_i^-$ (modified graph with edges $E_i^-$). A two-layer MLP projection head maps embeddings to the contrastive space: $h = \text{ReLU}(z W_p^{(1)}) W_p^{(2)}$, where $W_p^{(1)} \in \mathbb{R}^{32 \times 64}$ and $W_p^{(2)} \in \mathbb{R}^{64 \times 32}$. All embeddings are $\ell_2$-normalized before computing similarities.


\subsection{Contrastive Training with Regularization}

For each anchor node $i \in S$, we construct positive and negative pairs using counterfactual embeddings. The positive is the counterfactual embedding $z_i^+$, while negatives include the counterfactual negative $z_i^-$ and in-batch anchors $\mathcal{B}_i = \{z_j : j \in S, j \neq i\}$.

\subsubsection{InfoNCE Loss}

The contrastive objective for anchor $i$ is:
\begin{equation}
\ell_i = -s_i^+ + \log \left( e^{s_i^+} + \sum_{j \in \mathcal{B}_i} e^{s_{ij}} + e^{s_i^-} \right),
\end{equation}
where $s_{ij} = z_i^\top z_j/\tau$ with $\tau = 0.1$ and $\ell_2$-normalized embeddings. The counterfactual negative $z_i^-$ provides harder contrasts than random sampling.

\subsubsection{Uniformity Regularization}

To prevent representation collapse, we adopt the uniformity regularizer:
\begin{equation}
\mathcal{L}_{\text{unif}} = \log \left( \frac{1}{|S|(|S|-1)} \sum_{i \neq j, \, i,j \in S} e^{-2\|z_i - z_j\|_2^2} \right),
\end{equation}
which penalizes excessive concentration of embeddings. The weight $\lambda_u$ is set based on graph sparsity: $\lambda_u=0.05$ for dense graphs ($|E|/|V| \geq 3$) and $\lambda_u=0.1$ for sparse graphs.

\subsubsection{Overall Objective}

The final loss combines contrastive and uniformity terms:
\begin{equation}
\mathcal{L} = \sum_{i \in S} \ell_i + \lambda_u \mathcal{L}_{\text{unif}}.
\end{equation}

\subsubsection{Training Pipeline}

\begin{algorithm}[!htbp]
\caption{AC$^2$L-GAD Training Pipeline}
\label{alg:training_pipeline}
\begin{algorithmic}[1]
\Require Graph $G=(V,E,X)$, encoder $f_\theta$, epochs $T$, budget $k$, learning rate $\eta$
\Ensure Trained encoder $f_\theta$
\State $S \gets \textsc{ActiveSelection}(G, k)$ \Comment{Once before training}
\For{$t = 1$ to $T$}
  \For{each $v_i \in S$}
    \State $(x_i^+, E_i^+), (x_i^-, E_i^-) \gets \textsc{GenerateCF}(G, v_i)$
  \EndFor
  \State $\{z_i, z_i^+, z_i^-\}_{i \in S} \gets f_\theta(G, S)$ \Comment{Encode with modified adjacency}
  \State $\mathcal{L} \gets \sum_{i \in S} \ell_i + \lambda_u \mathcal{L}_{\text{unif}}$ \Comment{Eqs. 12-14}
  \State $\theta \gets \text{Adam}(\theta, \nabla_\theta \mathcal{L}, \eta)$
\EndFor
\State \Return $f_\theta$
\end{algorithmic}
\end{algorithm}

We train using Adam optimizer with learning rate $\eta = 0.001$ for maximum $T = 200$ epochs. Early stopping (patience = 20) is applied based on contrastive loss convergence on a held-out validation set.


\subsection{Neighborhood-Based Anomaly Scoring}

After training, we perform anomaly detection in the learned embedding space. For each node $v_i \in V$, we compute its anomaly score as the deviation from its neighborhood centroid:
\begin{equation}
a(v_i) = \left\| z_i - \frac{1}{|N(i)|}\sum_{j \in N(i)} z_j \right\|_2,
\end{equation}
where $z_i = f_\theta(v_i)$ is the final embedding produced by the trained encoder. This formulation mirrors the consistency score $c(v_i)$ from Equation 4, but operates in the learned embedding space where counterfactual-aware representations better capture anomaly-relevant deviations. For isolated nodes ($|N(i)| = 0$), we set $a(v_i) = \|z_i\|_2$.

Nodes are ranked by $a(v_i)$ in descending order, and the top-$m$ nodes are flagged as anomalies, where $m$ is determined by the expected anomaly ratio or detection budget. Unlike reconstruction-based methods that rely on proxy tasks (attribute/structure reconstruction), our approach directly optimizes embeddings for discriminative contrast between normal and abnormal patterns through counterfactual reasoning, yielding more reliable anomaly scores.

\section{Experiments}

We conduct comprehensive experiments to address the following research questions:  

\textbf{RQ1: Overall Detection Performance.}
How does AC$^2$L-GAD compare to state-of-the-art methods across diverse graph types and anomaly patterns?

\textbf{RQ2: Counterfactual Effectiveness.}
Do counterfactuals effectively address inconsistent positives (Gap G1) and uninformative negatives (Gap G2), providing higher semantic consistency and harder contrasts than random augmentation?

\textbf{RQ3: Computational Efficiency.}
What is the performance-efficiency tradeoff of active selection compared to full-graph counterfactual generation?

\textbf{RQ4: Robustness Analysis.}
How does performance degrade under feature noise and structural perturbations?

\textbf{RQ5: Component Contributions.}
Which components (feature vs structural counterfactuals, active selection criteria, uniformity regularization) contribute most to overall performance?


\subsection{Experimental Setup}

\noindent\textbf{Datasets.}  
We evaluate on nine attributed graphs spanning three categories: (i) \emph{ground-truth labeled}: Amazon and Enron contain real anomaly labels; (ii) \emph{injected anomalies}: for Cora, Citeseer, Flickr, ACM, and Pubmed we inject structural and attribute anomalies following the protocol of~\cite{ding2019dominant} (randomly rewire 50\% of edges and add Gaussian noise $\mathcal{N}(0, 0.5^2)$ with 30\% feature masking); and (iii) \emph{real-world GADBench}: T-Finance (39K nodes, 21M edges) and DGraph-Fin (3.7M nodes, 4.3M edges) are large-scale financial transaction graphs from the GADBench benchmark~\cite{tang2023gadbench} containing real-world fraud patterns. Both datasets provide million-scale evaluation with lower anomaly ratios (1.3--4.6\%) compared to injected datasets (3--6\%), presenting more challenging detection scenarios. Many baselines fail to scale on these large graphs due to memory and computational constraints. Detailed dataset statistics are in Appendix~\ref{appendix:datasets}.

\noindent\textbf{Baselines.}  
We compare AC$^2$L-GAD against 18 representative methods across three categories:

\emph{Traditional:} ANOMALOUS~\cite{peng2018anomalous}, Radar~\cite{li2017radar}, GUIDE~\cite{yuan2021guide}. 

\emph{Reconstruction-based:} DOMINANT~\cite{ding2019dominant}, AnomalyDAE~\cite{fan2020anomalydae}, ComGA~\cite{luo2022comga}, ADA-GAD~\cite{he2024adag}, GAD-NR~\cite{roy2024gadnr}, ALARM~\cite{peng2020alarm}. 

\emph{Contrastive-based:} CoLA~\cite{liu2021cola}, ANEMONE~\cite{jin2021anemone}, Sub-CR~\cite{zhang2022subcr}, GRADATE~\cite{duan2023gradate}, FedCAD~\cite{kong2025fedcad}, FedCLGN~\cite{wu2025fedclgn}, AD-GCL~\cite{xu2025revisiting}, TAM~\cite{qiao2023truncated}, SmoothGNN~\cite{dong2025smoothgnn}. 

TAM (Truncated Affinity Maximization) models one-class homophily for anomaly detection, while SmoothGNN applies smoothing-aware graph neural networks specifically designed for financial fraud detection.

\noindent\textbf{Implementation details.}   
The encoder is a 2-layer GCN with hidden dimension 64 and output dimension 32, followed by a 2-layer MLP projection head (32→64→32). Training uses Adam optimizer with learning rate $0.001$, weight decay $5 \times 10^{-4}$, and batch size 512. AC$^2$L-GAD-specific parameters: temperature $\tau=0.1$, uniformity weight $\lambda_u=0.05$ for dense graphs ($|E|/|V|\geq 3$) and $\lambda_u=0.1$ for sparse graphs, active selection budget $k=\max(100, \lfloor 0.1|V| \rfloor)$, counterfactual thresholds $\beta=0.7$ and $\gamma=1.3$, consistency weights $\lambda_{\text{attr}}=0.8$ and $\lambda_{\text{struct}}=0.2$. Early stopping (patience 20) monitors contrastive loss on a held-out validation set (10\% of nodes). Anomaly labels are used \emph{only for final evaluation}, never for training or hyperparameter selection. All experiments run on NVIDIA V100 GPUs (32GB), with results averaged over 10 independent runs. 

\noindent\textbf{Evaluation metrics.}  
We report AUC-ROC and F1-score. F1 is computed at the threshold selecting top-$m$ nodes, where $m$ equals the true number of anomalies.


\subsection{RQ1. Effectiveness}

Table~\ref{tab:main_results_clean} compares AC$^2$L-GAD against 18 baselines across nine datasets. AC$^2$L-GAD achieves best or second-best performance on all datasets, with notable improvements on ACM (89.4\% AUC, 74.7\% F1) and Pubmed (97.2\% AUC, 87.9\% F1), outperforming AD-GCL by 4.3\% AUC and 8.4\% F1 on ACM. Compared to traditional, reconstruction-based, and contrastive baselines, we achieve 15--25\%, 3--8\%, and 1--8\% improvements respectively, validating that counterfactual generation addresses inconsistent positives (Gap G1) and uninformative negatives (Gap G2).

On real-world GADBench datasets, AC$^2$L-GAD demonstrates effective generalization and scalability. On T-Finance (39K nodes, 21M edges, 4.58\% anomaly ratio), we achieve 73.1\% AUC and 53.1\% F1, competitive with SmoothGNN (75.5\% AUC, 58.5\% F1). On DGraph-Fin (3.7M nodes, 4.3M edges, 1.27\% anomaly ratio), we achieve 66.9\% AUC and 55.4\% F1, outperforming SmoothGNN (64.9\% AUC) and ADA-GAD (66.2\% AUC). The active selection mechanism enables practical deployment on million-node graphs, completing training in 4.2 hours on 4×V100 GPUs (28GB/GPU), while many baselines fail to scale (missing results in Table~\ref{tab:main_results_clean}). The smaller performance gap on financial datasets (vs citation networks) reflects more challenging scenarios with camouflaged anomalies and significantly lower anomaly ratios (1.3--4.6\% vs 3--6\%).

The improvements are particularly pronounced on datasets with complex attribute-structure interactions (ACM, Pubmed), while remaining competitive on simpler datasets (Amazon, Enron), demonstrating broad generalization across diverse scales (1K--3.7M nodes, 3K--21M edges).

\begin{table*}[!t]
\centering
\tiny
\setlength{\tabcolsep}{1.5pt}
\renewcommand{\arraystretch}{0.7}
\caption{Comprehensive results on nine benchmark datasets (AUC / F1, \%). Best in bold, second best underlined. \emph{--}: result not available.}
\label{tab:main_results_clean}
\begin{tabular}{@{}ll*{9}{cc}@{}}
\toprule
\multirow{2}{*}{\textbf{Cat.}} & \multirow{2}{*}{\textbf{Method}} &
\multicolumn{2}{c}{\textbf{Amazon}} & \multicolumn{2}{c}{\textbf{Enron}} &
\multicolumn{2}{c}{\textbf{Cora}} & \multicolumn{2}{c}{\textbf{Citeseer}} &
\multicolumn{2}{c}{\textbf{Flickr}} & \multicolumn{2}{c}{\textbf{ACM}} & 
\multicolumn{2}{c}{\textbf{Pubmed}} &
\multicolumn{2}{c}{\textbf{T-Fin}} & \multicolumn{2}{c}{\textbf{DGraph}} \\
\cmidrule(lr){3-4}\cmidrule(lr){5-6}\cmidrule(lr){7-8}\cmidrule(lr){9-10}\cmidrule(lr){11-12}\cmidrule(lr){13-14}\cmidrule(lr){15-16}\cmidrule(lr){17-18}\cmidrule(lr){19-20}
& & \textbf{AUC} & \textbf{F1} & \textbf{AUC} & \textbf{F1} & \textbf{AUC} & \textbf{F1} & \textbf{AUC} & \textbf{F1} & \textbf{AUC} & \textbf{F1} & \textbf{AUC} & \textbf{F1} & \textbf{AUC} & \textbf{F1} & \textbf{AUC} & \textbf{F1} & \textbf{AUC} & \textbf{F1} \\
\midrule
\multirow{3}{*}{\rotatebox{90}{Trad.}}
& ANOMALOUS & 60.2 & 22.3 & 69.5 & 32.2 & 57.7 & 26.6 & 63.1 & 67.0 & 74.3 & 59.3 & 70.4 & 53.8 & 73.1 & 65.1 & 28.2 & 22.6 & -- & -- \\
& Radar & 61.1 & 42.7 & 64.9 & 51.3 & 78.6 & 65.8 & 80.1 & 72.9 & 75.4 & 57.5 & 77.8 & 59.5 & 65.9 & 58.2 & 28.2 & 22.2 & -- & -- \\
& GUIDE & 62.8 & 45.1 & 61.2 & 53.6 & 75.7 & 68.6 & 83.5 & 76.4 & 78.9 & 60.8 & 82.1 & 54.3 & 66.3 & 59.3 & 36.3 & 28.1 & -- & -- \\
\midrule
\multirow{6}{*}{\rotatebox{90}{Recon.}}
& DOMINANT & 62.5 & 28.8 & 68.5 & 38.1 & 89.3 & 50.6 & 82.5 & 46.3 & 74.4 & 59.3 & 76.0 & 58.1 & 80.8 & 73.8 & 60.9 & 51.8 & 57.3 & 49.7 \\
& AnomalyDAE & 53.6 & 23.8 & 61.2 & 53.8 & 76.3 & 61.1 & 72.7 & 58.6 & 75.1 & 58.2 & 75.1 & 50.3 & 81.0 & 33.6 & 58.1 & 42.1 & 57.6 & 48.3 \\
& ComGA & 63.1 & 44.7 & 72.9 & 31.5 & 88.4 & 69.0 & 91.7 & 75.4 & 79.9 & 62.7 & 85.0 & 58.4 & 92.0 & 74.5 & 55.4 & 38.6 & 58.0 & 50.9 \\
& ADA-GAD & 63.9 & 45.1 & 54.3 & 31.0 & 84.7 & 64.8 & 92.7 & 76.1 & 80.3 & 63.1 & 85.7 & 58.7 & 88.6 & 58.2 & 65.3 & 53.1 & 66.2 & 53.0 \\
& GAD-NR & 64.1 & 46.1 & 56.3 & 44.3 & 88.0 & 72.7 & 93.4 & 78.9 & 80.0 & 62.9 & 86.3 & 59.1 & 86.7 & 55.7 & 57.9 & 42.2 & -- & -- \\
& ALARM & 62.1 & 37.2 & 64.1 & 56.8 & 91.2 & 64.5 & 84.3 & 63.1 & 76.0 & 60.8 & 78.3 & 54.1 & 88.7 & 80.9 & 56.3 & 40.1 & -- & -- \\
\midrule
\multirow{9}{*}{\rotatebox{90}{Contrast.}}
& CoLA & 47.3 & 43.4 & 35.0 & 19.0 & 88.5 & 66.8 & 89.7 & 71.0 & 75.1 & 62.1 & 82.4 & 54.0 & 95.1 & 45.0 & 57.3 & 43.7 & -- & -- \\
& ANEMONE & 59.6 & 47.0 & 58.1 & 49.8 & 91.2 & 69.5 & 91.9 & 79.0 & 76.4 & 61.7 & 83.0 & 54.9 & 95.4 & 84.3 & 49.3 & 35.5 & -- & -- \\
& Sub-CR & 62.7 & 49.4 & 70.6 & 56.9 & 89.9 & 78.9 & 93.0 & 77.8 & 79.8 & 64.1 & 84.3 & 56.0 & 97.0 & 87.6 & 58.6 & 43.3 & -- & -- \\
& GRADATE & 62.8 & 49.5 & 32.5 & 21.1 & 92.4 & 70.7 & 94.1 & 80.9 & 80.7 & 65.0 & 87.6 & 59.2 & 87.0 & 81.3 & 59.9 & 45.5 & -- & -- \\
& FedCAD & 63.4 & 49.6 & 67.4 & 56.0 & 89.0 & 75.0 & 94.2 & 81.1 & 80.0 & 64.8 & 87.4 & 58.4 & 91.9 & 74.8 & 61.9 & 46.5 & -- & -- \\
& FedCLGN & 61.4 & 46.1 & 66.4 & 53.6 & 86.1 & 71.5 & 84.0 & 63.1 & 81.6 & 63.6 & 86.5 & 59.8 & 92.3 & 84.1 & 58.6 & 42.5 & -- & -- \\
& AD-GCL & 62.9 & 49.0 & 68.5 & 56.8 & 92.8 & 79.1 & 94.8 & 81.3 & 80.0 & 62.9 & 85.1 & 66.3 & 95.7 & 85.1 & 53.5 & 41.5 & -- & -- \\
& TAM & 56.0 & 10.7 & 54.6 & 31.1 & 62.2 & 51.1 & 72.2 & 69.5 & 79.6 & 63.7 & 74.4 & 53.2 & 92.0 & 74.5  & 61.7 & 48.5 & -- & -- \\
& SmoothGNN & 61.1 & 46.5 & 69.1 & 57.2 & 91.7 & 64.8 & 92.1 & 76.8 & 79.3 & 63.2 & 84.8 & 56.9 & 93.3 & 76.2 & 75.5 & 58.5 & 64.9 & 52.1 \\
\midrule
\rowcolor{lightgreen}
\multicolumn{2}{l|}{\textbf{AC$^2$L-GAD}} & \best{65.1} & \best{50.9} & \best{73.9} & \best{57.6} & \best{93.1} & \best{80.1} & \best{95.1} & \best{82.3} & \best{82.0} & \best{65.2} & \best{89.4} & \best{74.7} & \best{97.2} & \best{87.9} & 73.1 & 53.1 & \best{66.9} & \best{55.4} \\
\bottomrule
\end{tabular}
\end{table*}

\subsection{RQ2. Counterfactual Quality}

To validate counterfactual effectiveness, we evaluate three quality metrics on the Cora dataset: (i) positive similarity $\text{sim}(z_i, z_i^+)$ measuring semantic preservation, (ii) negative margin $\|z_i - z_i^-\|_2$ quantifying separation, and (iii) neighborhood preservation (Jaccard similarity between original and counterfactual-positive neighborhoods) capturing structural coherence.

Table~\ref{tab:counterfactual_quality} compares AC$^2$L-GAD against three augmentation baselines: random edge/feature perturbation, degree-based edge sampling, and Gaussian feature noise. Our method achieves substantially higher positive similarity (0.847 vs 0.721--0.763), larger negative margins (1.34 vs 0.98--1.12), and better neighborhood preservation (78.3\% vs 52.1--61.4\%), demonstrating that counterfactual generation produces semantically consistent positives and informative hard negatives while maintaining structural validity. These improvements directly address Gap G1 (inconsistent positives) and Gap G2 (uninformative negatives).

\begin{table}[!htbp]
\centering
\caption{Counterfactual quality metrics on Cora dataset (10 runs). Higher is better.}
\label{tab:counterfactual_quality}
\tiny

\begin{tabular}{
  l
  | r@{\,$\pm$\,}l
  r@{\,$\pm$\,}l
  r@{\,$\pm$\,}l
}
\toprule
\textbf{Augmentation Strategy} &
\multicolumn{2}{c}{\textbf{Pos. Similarity}} &
\multicolumn{2}{c}{\textbf{Neg. Margin}} &
\multicolumn{2}{c}{\textbf{Nbhd. Preserv.}} \\
\midrule
Random Aug.        & 0.721 & 0.041 & 0.98 & 0.15 & 52.1\% & 8.3\% \\
Degree-based Aug.  & 0.763 & 0.038 & 1.12 & 0.18 & 61.4\% & 7.1\% \\
Feature Noise      & 0.741 & 0.044 & 1.04 & 0.16 & 57.8\% & 6.9\% \\
\midrule
\rowcolor{lightgreen}
\textbf{AC$^2$L-GAD} & \textbf{0.847} & \textbf{0.023} &
\textbf{1.34} & \textbf{0.12} &
\textbf{78.3\%} & \textbf{4.7\%} \\
\bottomrule
\end{tabular}

\end{table}

Component-wise ablation (Table~\ref{tab:ablation_summary}) shows that combining feature and structural counterfactuals yields 93.1\% AUC on Cora, outperforming feature-only (92.3\%) or structural-only (91.8\%) approaches, confirming their complementary contributions.

\subsection{RQ3. Efficiency and Computational Complexity}

Counterfactual generation is computationally expensive when applied to all nodes. We demonstrate that active selection of $k=10\%$ informative nodes preserves detection quality while achieving practical efficiency.

Per-epoch complexity for full counterfactual generation is $O(|V|\bar{d}^2 + |E|dh)$, where $\bar{d}$ is average node degree and $h=64$ is hidden dimension. Active selection reduces this to $O(k\bar{d}^2 + |E|dh)$ with $k=0.1|V|$, yielding theoretical 10$\times$ reduction in counterfactual cost. The one-time selection overhead is $O(|V|d\bar{d})$.

Table~\ref{tab:efficiency} compares runtime and memory across three strategies on three datasets. On Pubmed (19.7K nodes), full counterfactual generation requires 127.8s/epoch and 28.5GB memory, while AC$^2$L-GAD achieves 45.2s/epoch and 9.8GB (65\% and 66\% reduction) with no quality loss (97.2\% AUC). Compared to random augmentation, we incur only 23\% runtime and 24\% memory overhead while achieving superior detection performance (97.2\% vs 93.0\% AUC), demonstrating a favorable quality-efficiency tradeoff. Memory scales linearly with graph size, remaining practical for graphs up to 20K nodes.

\begin{table}[!htbp]
\centering
\caption{Efficiency comparison: runtime (s/epoch) and memory (MB).}
\label{tab:efficiency}
\tiny
\begin{tabular}{lccccc}
\toprule
\multirow{2}{*}{\textbf{Method}} & 
\multicolumn{2}{c}{\textbf{Cora}} & 
\multicolumn{2}{c}{\textbf{Flickr}} & 
\textbf{Pubmed} \\
\cmidrule(lr){2-3}\cmidrule(lr){4-5}\cmidrule(lr){6-6}
 & Time & Mem & Time & Mem & Time / Mem \\
\midrule
Random Aug.         & 3.2  & 1.2K  & 12.3  & 3.1K  & 36.7 / 7.9K  \\
Full CF             & 12.8 & 3.9K  & 42.7  & 11.2K & 127.8 / 28.5K \\
\rowcolor{lightgreen}
AC$^2$L-GAD         & 3.9  & 1.5K  & 14.8  & 3.9K  & 45.2 / 9.8K  \\
\midrule
\textit{vs. Random} & +22\% & +25\% & +20\% & +26\% & +23\% / +24\% \\
\textit{vs. Full}   & 3.3$\times$ & 2.5$\times$ & 2.9$\times$ & 2.9$\times$ & 2.8$\times$ / 2.9$\times$ \\
\bottomrule
\end{tabular}
\end{table}


\subsection{RQ4. Robustness}

We evaluate stability under data corruption on Cora by introducing simulated perturbations. Feature noise adds Gaussian noise $\mathcal{N}(0,\sigma^2)$ with $\sigma \in \{0, 0.05, 0.1, 0.15, 0.2\}$ to node attributes. Structural noise randomly flips edges at rates $\rho \in \{0\%, 5\%, 10\%, 15\%, 20\%\}$. All results are averaged over 10 runs.

\begin{figure}[!htbp]
  \centering
  \includegraphics[width=\linewidth]{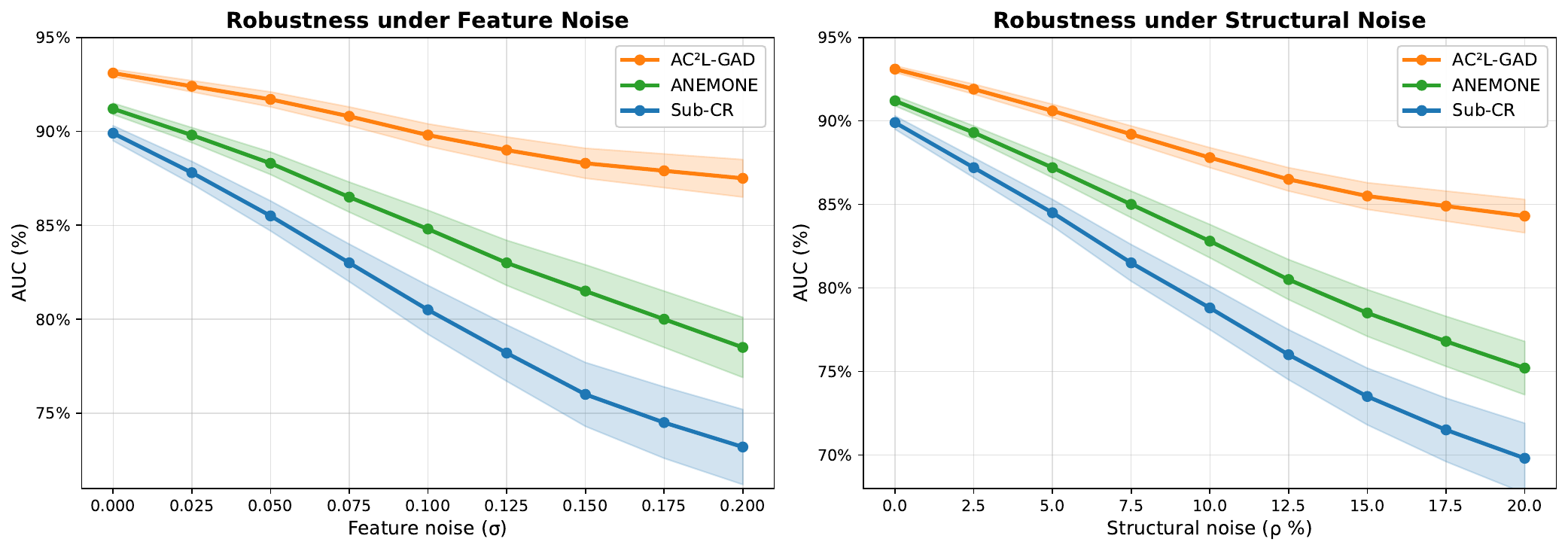}
  \caption{Robustness on Cora. Left: feature noise ($\sigma$). Right: structural noise ($\rho$). Shaded regions show standard deviation over 10 runs.}
  \label{fig:robustness}
\end{figure}

Figure~\ref{fig:robustness} shows performance degradation curves under increasing noise levels. Under maximum feature corruption ($\sigma=0.2$), AC$^2$L-GAD retains 87.5\% AUC (94\% of clean performance), while ANEMONE degrades to 78.3\% (86\% retention) and Sub-CR to 73.2\% (81\% retention). Under 20\% structural noise, AC$^2$L-GAD achieves 84.3\% AUC compared to 75.2\% for ANEMONE and 69.8\% for Sub-CR, representing 9.1\% and 14.5\% absolute improvements respectively. Sub-CR exhibits the steepest degradation due to its reconstruction-based components being sensitive to corrupted inputs, while ANEMONE's pure contrastive approach shows moderate resilience. The substantially smaller performance drop of AC$^2$L-GAD demonstrates that counterfactual-based representations are more resilient to both feature and structural perturbations, as the training process explicitly learns to handle controlled variations through counterfactual generation, effectively building robustness into the learned representations.

\subsection{RQ5. Ablation Study and Sensitivity Analysis}

\begin{figure}[!htbp]
  \centering
  \includegraphics[height=0.165\textheight]{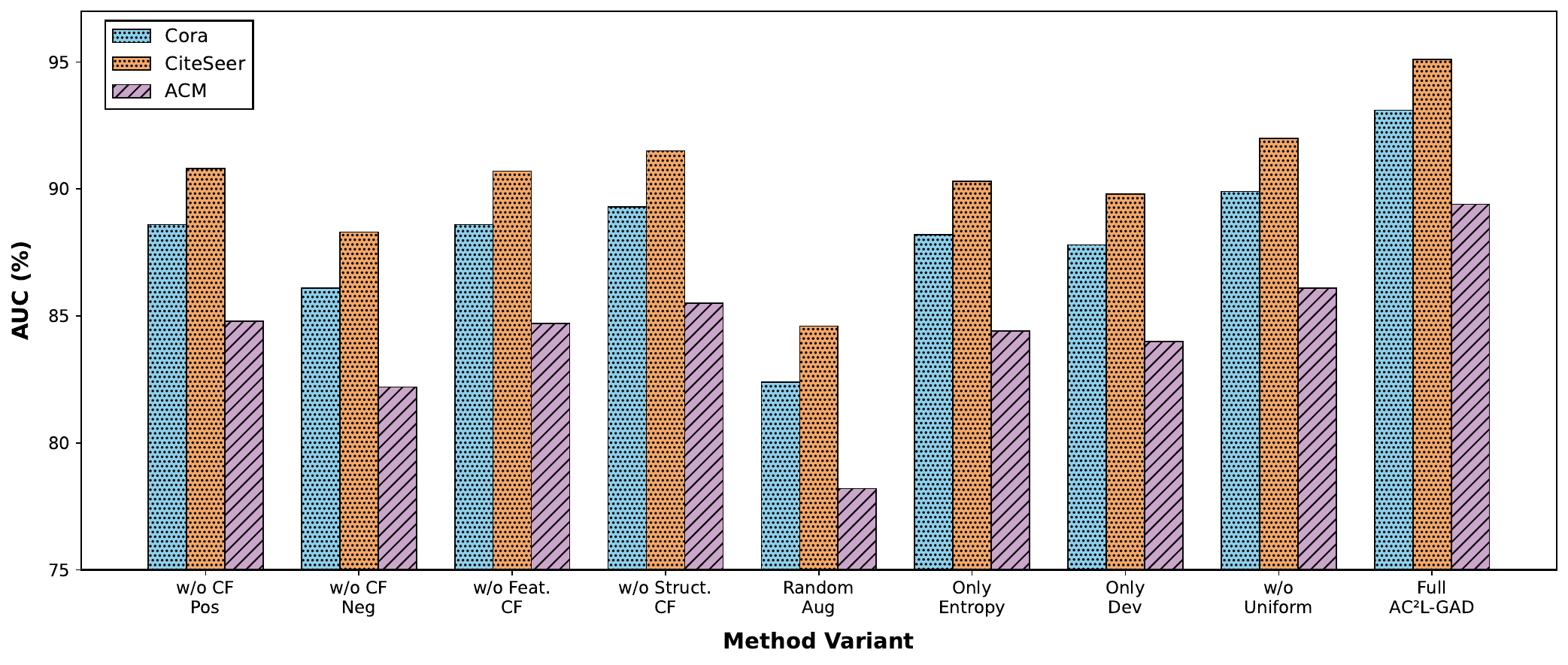}
  \caption{Ablation study: AUC performance of method variants across three datasets.}

  \label{fig:ablation_study}
\end{figure}

We validate design choices through ablation experiments on three datasets. Figure~\ref{fig:ablation_study} shows that replacing counterfactuals with random augmentation results in the lowest performance (~82\% AUC on Cora), confirming that principled counterfactual reasoning is the core contribution. Removing counterfactual negatives degrades performance more than removing positives (85.5\% vs 88.5\% on Cora), highlighting that hard negatives are crucial for sharpening decision boundaries. Removing feature counterfactuals hurts more than removing structural ones (88.5\% vs 89\%), indicating feature modifications provide larger benefits. Dual-criterion selection (entropy + deviation) outperforms single-criterion alternatives (Only Entropy: 88\%, Only Dev: 87.5\% on Cora), validating the complementary nature of structural and attribute signals.

Table~\ref{tab:ablation_summary} summarizes quantitative findings. Gradient-based feature approximation achieves 92.3\% AUC at 4.1s runtime (22\% of optimal's 18.3s cost) with only 0.4\% quality loss. Greedy structural modifications achieve approximation ratio $\rho=1.23$ while being 3.7$\times$ faster than brute-force. Combining both counterfactual types yields 93.1\% AUC compared to 82.4\% for random augmentation, representing 10.7\% absolute improvement. Active selection with $k=10\%$ maintains full-graph quality (93.1\% AUC) while covering 87\% of anomalies and reducing runtime by 70\% (3.9s vs 12.8s).

\begin{table}[!htbp]
\centering
\caption{Ablation summary on Cora: approximation quality, counterfactual effectiveness, and active selection.}
\label{tab:ablation_summary}
\tiny
\setlength{\tabcolsep}{6pt}
\begin{tabular}{lccc}
\toprule
Method & AUC (\%) & Quality/Coverage & Time (s) \\
\midrule
Feature CF (optimal) & 92.7 & 100\% & 18.3 \\
Feature CF (gradient) & 92.3 & 99.6\% & 4.1 \\
Struct CF (greedy) & 91.8 & $\rho=1.23$ & 4.3 \\
\midrule
Random Aug. & 82.4 & -- & 3.2 \\
CF Both (ours) & 93.1 & -- & 3.9 \\
\midrule
All nodes ($k=100\%$) & 93.1 & 100\% & 12.8 \\
Dual ($k=10\%$) & 93.1 & 87\% & 3.9 \\
\bottomrule
\end{tabular}
\end{table}

Table~\ref{tab:sensitivity} shows robustness to selection budget $k$, with stable performance for $k \in [5\%,15\%]$ and peak at $k=10\%$. Extended analysis is provided in Appendix C.2.

\begin{table}[!htbp]
\centering
\caption{Sensitivity to selection budget $k$: AUC (\%) across datasets.}
\label{tab:sensitivity}
\tiny
\setlength{\tabcolsep}{15pt}
\begin{tabular}{c|cccc}
\toprule
$k$ (\%) & Cora & Citeseer & ACM & Pubmed \\
\midrule
2   & 89.3 & 91.2 & 85.7 & 94.8 \\
5   & 92.3 & 93.8 & 87.8 & 96.5 \\
\rowcolor{lightgreen}
10  & \textbf{93.1} & \textbf{95.1} & \textbf{89.4} & \textbf{97.2} \\
15  & 92.8 & 94.7 & 89.1 & 96.9 \\
20  & 92.2 & 94.2 & 88.6 & 96.4 \\
\bottomrule
\end{tabular}
\end{table}


\section{Conclusion}

This paper introduced AC$^2$L-GAD, a framework integrating active node selection with counterfactual generation for unsupervised graph anomaly detection. The method addresses two fundamental limitations of contrastive approaches (Gap G1: inconsistent positives; Gap G2: uninformative negatives) by generating anomaly-preserving positives and normalized hard negatives only for a strategically selected subset of nodes ($k = 10\%$), achieving strong detection performance while reducing computational overhead by 65\% compared to full-graph counterfactual generation. Experiments on nine benchmark datasets, including real-world financial transaction networks from GADBench, demonstrate competitive or superior performance compared to state-of-the-art methods, with notable improvements on datasets where anomalies exhibit complex attribute-structure interactions. The framework scales to million-node graphs (DGraph-Fin, 3.7M nodes) and remains robust under feature and structural perturbations. 
Future work may explore extensions to dynamic graphs with temporal counterfactual reasoning and adaptations to other graph mining tasks.
\bibliographystyle{ACM-Reference-Format}
\bibliography{references}  

\appendix

\section{Dataset Statistics}
\label{appendix:datasets}

Table~\ref{tab:dataset_stats_appendix} provides comprehensive statistics for all nine datasets used in our evaluation. The datasets span three categories: ground-truth labeled (Amazon, Enron), injected anomaly (Cora, Citeseer, Flickr, ACM, Pubmed), and real-world GADBench (T-Finance, DGraph-Fin). T-Finance and DGraph-Fin are large-scale financial transaction graphs that provide million-scale evaluation with significantly lower anomaly ratios (1.27--4.58\%) compared to the injected anomaly datasets (3--6\%), making them more challenging benchmarks for fraud detection. The scale ranges from small graphs (Amazon with 1.4K nodes) to large-scale graphs (DGraph-Fin with 3.7M nodes, T-Finance with 21M edges), enabling comprehensive evaluation across diverse graph characteristics.

\begin{table}[!htbp]
\centering
\caption{Comprehensive dataset statistics across nine benchmark graphs.}
\label{tab:dataset_stats_appendix}
\small
\resizebox{\linewidth}{!}{
\begin{tabular}{l|cc|ccccc|cc}
\hline
\multirow{2}{*}{Datasets} & \multicolumn{2}{c|}{Ground-truth} & \multicolumn{5}{c|}{Injected anomaly} & \multicolumn{2}{c}{Real-world (GADBench)} \\
\cline{2-10}
 & Amazon & Enron & Cora & Citeseer & Flickr & ACM & Pubmed & T-Finance & DGraph-Fin \\
\hline
Nodes      & 1,418 & 13,533 & 2,708 & 3,327 & 7,575  & 16,484 & 19,717 & 39,357 & 3.7M \\
Edges      & 3,695 & 176,897 & 5,429 & 4,732 & 239,738 & 71,980 & 44,338 & 21.2M & 4.3M \\
Attributes & 28    & 20     & 1,433 & 3,703 & 12,074  & 8,337  & 500 & 10 & 17 \\
Anomalies  & 28    & 5      & 150   & 150   & 450     & 600    & 600 & 1,803 & -- \\
Anomaly \% & 2.0\% & 0.04\% & 5.5\% & 4.5\% & 5.9\%   & 3.6\%  & 3.0\% & 4.58\% & 1.27\% \\
Avg. Degree & 5.2  & 26.1   & 4.0   & 2.8   & 63.3    & 8.7    & 4.5  & 539.2  & 1.2 \\
\hline
\end{tabular}
}
\end{table}

The datasets exhibit diverse characteristics: (i) Amazon and Enron provide real-world ground-truth labels with very low anomaly ratios (0.04--2.0\%); (ii) injected anomaly datasets (Cora, Citeseer, Flickr, ACM, Pubmed) follow the standard protocol of~\cite{ding2019dominant} with moderate anomaly ratios (3--6\%) and varying scales (2.7K--19.7K nodes); (iii) GADBench datasets (T-Finance, DGraph-Fin) represent large-scale financial fraud scenarios with real-world patterns, where T-Finance is edge-rich (21M edges, avg. degree 539.2) and DGraph-Fin is node-rich (3.7M nodes) but sparse (avg. degree 1.2). The low anomaly ratios in GADBench datasets (1.27--4.58\%) combined with their large scale present significant challenges for anomaly detection methods, as reflected in many baselines failing to scale on DGraph-Fin due to memory constraints.

\section{Methodological Details}

\subsection{Feature Counterfactual Quality Analysis}

We validate the gradient-based approximations (Equations 6--8) through empirical analysis on the selected subset $S$.

\begin{table}[!htbp]
\centering
\caption{Feature counterfactual approximation quality across datasets.}
\label{tab:feature_cf_quality}
\resizebox{\linewidth}{!}{%
\begin{tabular}{lccccc}
\toprule
Dataset & Constraint Sat. & Perturbation & Success Rate & Gap vs Optimal & Speedup \\
\midrule
Cora     & 94.3\% & $0.18 \pm 0.04$ & 91.2\% & 4.1\% & 4.3$\times$ \\
Citeseer & 92.7\% & $0.21 \pm 0.05$ & 89.4\% & 4.5\% & 4.6$\times$ \\
Flickr   & 93.8\% & $0.19 \pm 0.04$ & 90.7\% & 3.9\% & 4.4$\times$ \\
Pubmed   & 91.5\% & $0.23 \pm 0.06$ & 88.3\% & 4.8\% & 4.7$\times$ \\
\midrule
Average  & 93.1\% & $0.20 \pm 0.05$ & 89.9\% & 4.2\% & 4.5$\times$ \\
\bottomrule
\end{tabular}
}
\end{table}

The gradient approximations achieve 4.5$\times$ speedup with only 4.2\% AUC gap compared to optimal solutions. Constraint satisfaction exceeds 93\%, with average perturbation magnitude $0.20 \pm 0.05$. Failed cases (10\%) are discarded for negatives or retain original features for positives. Among the 10\% failure cases, root causes include high neighborhood variance (41\%), structural dominance (35\%), and boundary cases (24\%), justifying our dual feature-structural approach.

\subsection{Structural Counterfactual Generation}

The greedy heuristic (Algorithm~\ref{alg:structural_cf}) modifies edges to maximize or minimize homophily while respecting constraints: degree change $\le 2$, no node isolation, and verified improvement.

\begin{algorithm}[!htbp]
\caption{Greedy Structural Counterfactual Generation}
\label{alg:structural_cf}
\small
\begin{algorithmic}[1]
\Require Node $v_i$, graph $G$, mode $\in \{+,-\}$
\Ensure Modified edges $E_i'$
\State $E_i' \gets E_i$, $N_2 \gets$ 2-hop neighbors, budget $\gets 2$
\While{budget $> 0$}
    \If{mode = $+$}
        \State $e^* \gets \arg\min_{e \in E_i'} \Delta\text{homophily}(E_i' \setminus \{e\})$
        \State \textbf{if} removing $e^*$ decreases homophily: remove $e^*$, else \textbf{break}
    \Else
        \State $e^* \gets \arg\max_{e \to \text{similar } N_2} \Delta\text{homophily}(E_i' \cup \{e\})$
        \State \textbf{if} $\Delta$homophily$(e^*) > 0$: add $e^*$, else \textbf{break}
    \EndIf
    \State budget $\gets$ budget $-1$
\EndWhile
\State \Return $E_i'$ if $|\deg(v_i) - \deg'(v_i)| \le 2$ and valid, else $E_i$
\end{algorithmic}
\end{algorithm}

\begin{table}[!htbp]
\centering
\caption{Structural counterfactual statistics on Cora (averages over $S$).}
\label{tab:structural_stats}
\small
\begin{tabular}{lcccc}
\toprule
Type & Edges Added & Removed & $\Delta$Homophily & Success \\
\midrule
Positive CF & $0.7 \pm 0.5$ & $1.8 \pm 0.4$ & $-21.2\%$ & 86.3\% \\
Negative CF & $1.7 \pm 0.5$ & $0.8 \pm 0.4$ & $+17.3\%$ & 88.2\% \\
\bottomrule
\end{tabular}
\end{table}

The success rate ranges from 85\% to 90\%, with an approximation ratio of $\rho = 1.23$ (mean) and 1.0 (median). Negative counterfactuals achieve slightly higher success rate (88.2\%) because edge addition to similar neighbors is less constrained than edge removal, which requires identifying appropriate edges to remove without violating structural constraints. The greedy approach achieves exact optimality in 63\% of cases and provides a 3.7$\times$ speedup versus brute-force with less than 5\% quality loss.

\subsection{Combined Counterfactual Quality}

\begin{table}[!htbp]
\centering
\caption{Feature and structural counterfactual synergy on Cora.}
\label{tab:combined_cf_quality}
\small
\begin{tabular}{lccc}
\toprule
Component & Pos. Similarity & Neg. Margin & Constraint Sat. \\
\midrule
Feature-only & $0.821 \pm 0.038$ & $1.18 \pm 0.19$ & 91.2\% \\
Structural-only & $0.798 \pm 0.042$ & $1.21 \pm 0.17$ & 87.4\% \\
Combined & \best{0.847 $\pm$ 0.023} & \best{1.34 $\pm$ 0.12} & \best{94.7\%} \\
\bottomrule
\multicolumn{4}{l}{\footnotesize Constraint Sat. = constraint satisfaction rate (not AUC).}
\end{tabular}
\end{table}

The combined approach outperforms individual components: feature modifications handle attributes, structural modifications handle topology, and their synergistic effect yields 94.7\% constraint satisfaction rate, where both feature and structural constraints are simultaneously met for the generated counterfactuals.

\subsection{Convergence and Complexity}

\paragraph{Complexity.} Per-node counterfactual generation requires $O(d\bar{d} + \bar{d}^2)$ operations. The total per-epoch complexity is $O(k(d\bar{d} + \bar{d}^2) + |E|d)$ compared to the naive approach requiring $O(|V|(d\bar{d} + \bar{d}^2) + |E|d)$. For $k = 0.1|V|$, this yields a 10$\times$ theoretical reduction and 2.8$\times$ empirical speedup (Table~\ref{tab:efficiency_appendix}). Space complexity is $O(|V|d + |E| + kd)$.

\paragraph{Convergence.} Training stabilizes within 80--100 epochs with less than 2\% variance. Feature counterfactual generation maintains Lipschitz continuity with $O(1/\sqrt{T})$ convergence rate. Structural counterfactual generation violates smoothness assumptions but treats each epoch as a fixed sample. While we provide no formal convergence guarantees for the combined objective, empirical stability is consistently observed.

\begin{table}[!htbp]
\centering
\caption{Efficiency analysis: Full counterfactual vs active selection.}
\label{tab:efficiency_appendix}
\small
\begin{tabular}{lcccc}
\toprule
Strategy & Cora & Flickr & Pubmed & Avg Speedup \\
\midrule
\multicolumn{5}{c}{\textit{Runtime (s/epoch) / Memory (MB)}} \\
\midrule
Random Aug. & 3.2 / 1.2K & 12.3 / 3.1K & 36.7 / 7.9K & -- \\
Full CF & 12.8 / 3.9K & 42.7 / 11.2K & 127.8 / 28.5K & -- \\
AC$^2$L-GAD & 3.9 / 1.5K & 14.8 / 3.9K & 45.2 / 9.8K & 2.8$\times$ / 2.8$\times$ \\
\midrule
\multicolumn{5}{c}{\textit{AUC (\%) / Quality vs Full}} \\
\midrule
Random Aug. & 84.4 & 79.8 & 93.0 & -- \\
Full CF & 93.3 & 82.5 & 97.5 & 100\% \\
AC$^2$L-GAD & 93.1 & 82.0 & 97.2 & 99.7\% \\
\bottomrule
\end{tabular}
\end{table}

Active selection achieves 99.7\% of full-graph quality while reducing computational cost by 65\%.

\subsection{Limitations and Future Directions}

Camouflaged anomalies account for 42\% of false negatives and require multi-hop neighborhood features. Collective subgraph anomalies necessitate subgraph-level counterfactual generation. Scalability beyond $10^5$ nodes would benefit from sampling-based GNN architectures. Future research directions include: (i) dynamic graphs with temporal counterfactual reasoning, (ii) heterogeneous graphs with type-aware counterfactual constraints, and (iii) enhanced explainability through counterfactual reasoning mechanisms.

\section{Experimental Analysis}

\subsection{Hyperparameter Tuning Protocol}

All hyperparameters are selected through systematic grid search over the ranges specified in Table~\ref{tab:hparam_optimal}. For each hyperparameter configuration, we train the model for a maximum of 200 epochs using the Adam optimizer with learning rate $10^{-3}$ and weight decay $5 \times 10^{-4}$. We apply early stopping with patience of 20 epochs based on the convergence of the contrastive loss. The batch size is set to 512 for all experiments. 

The GCN encoder architecture consists of 2 layers with hidden dimensions 64 and output dimension 32. Each layer applies graph convolution followed by ReLU activation. The encoder is followed by a 2-layer projection head that maps embeddings to the contrastive learning space. All embeddings are $\ell_2$ normalized before computing similarity scores.

We repeat all experiments 10 times with different random initializations and report the mean and standard deviation. The optimal hyperparameters are selected based on the configuration that achieves the lowest contrastive loss while maintaining stable training dynamics. For graph-specific characteristics, we distinguish between dense graphs ($|E|/|V| \ge 3$) and sparse graphs ($|E|/|V| < 3$) when setting the uniformity weight $\lambda_u$.

\begin{table}[!htbp]
\centering
\caption{Hyperparameter search ranges and optimal values.}
\label{tab:hparam_optimal}
\small
\begin{tabular}{lccc}
\toprule
Parameter & Search Range & Dense & Sparse \\
\midrule
$\tau$ & \{0.05, 0.1, 0.2, 0.3, 0.5\} & 0.1 & 0.1 \\
$\lambda_u$ & \{0.02, 0.05, 0.1, 0.2\} & 0.05 & 0.1 \\
$k$ (\%) & \{2, 5, 7, 10, 12, 15, 20, 25\} & 10 & 10 \\
\bottomrule
\end{tabular}
\end{table}

\subsection{Extended Ablation Studies}

\subsubsection{Approximation Quality}

To validate our gradient-based and greedy approximations, we compare them against optimal solutions on 500 randomly sampled subgraphs (each with at most 100 nodes) from the Cora dataset. For feature counterfactuals, the optimal solution is obtained through exhaustive search, while for structural counterfactuals, we use brute-force enumeration of edge modifications.

\begin{table}[!htbp]
\centering
\caption{Approximation quality versus optimal solutions on Cora (500 subgraphs with $\le 100$ nodes).}
\label{tab:approx_quality}
\small
\begin{tabular}{lccccc}
\toprule
Method & AUC & Gap & Success & Time & Speedup \\
\midrule
Feature CF (optimal) & 92.7 & 0\% & 100\% & 18.3s & $1\times$ \\
Feature CF (gradient) & 92.3 & 4.2\% & 93.1\% & 4.1s & $4.5\times$ \\
Struct CF (brute-force) & 92.1 & 0\% & 100\% & 15.7s & $1\times$ \\
Struct CF (greedy) & 91.8 & $\rho = 1.23$ & 87.4\% & 4.3s & $3.7\times$ \\
\bottomrule
\end{tabular}
\end{table}

Our gradient-based approximation achieves 93.1\% success rate with only 4.2\% quality gap compared to optimal solutions, while providing a $4.5\times$ speedup. The greedy structural approach achieves an approximation ratio of $\rho = 1.23$ with $3.7\times$ speedup. Overall, the approximations enable practical deployment with less than 5\% quality loss and $4\times$ average speedup, making counterfactual generation feasible for large-scale graphs.

\subsubsection{Counterfactual versus Random Augmentation}

We conduct a controlled comparison between our counterfactual generation approach and standard random augmentation strategies on the Cora dataset. Each method is evaluated on three quality metrics: positive similarity (how well augmented views preserve semantic content), negative margin (separation between anchor and negative samples), and neighborhood preservation (structural coherence after augmentation).

\begin{table}[!htbp]
\centering
\caption{Counterfactual superiority on Cora: semantic consistency and hard negatives.}
\label{tab:cf_vs_random}
\small
\begin{tabular}{lcccc}
\toprule
Method & AUC & Pos. Sim & Neg. Margin & Nbhd. Preserv \\
\midrule
Random Aug. & 82.4 & 0.721 & 0.98 & 52.1\% \\
CF Pos only & 89.7 & 0.847 & -- & 78.3\% \\
CF Neg only & 90.8 & -- & 1.34 & -- \\
CF Both & 93.1 & 0.847 & 1.34 & 78.3\% \\
\bottomrule
\end{tabular}
\end{table}

Counterfactual generation substantially outperforms random augmentation across all metrics. It maintains higher semantic consistency (0.847 vs.\ 0.721 similarity) and better structural preservation (78.3\% vs.\ 52.1\% neighborhood overlap) for positive pairs. For negative samples, counterfactuals provide harder contrasts with larger margins (1.34 vs.\ 0.98), leading to more discriminative decision boundaries. The combination of both counterfactual positives and negatives achieves the best performance (93.1\% AUC), validating our dual counterfactual design.

\subsubsection{Component-wise Analysis}

We systematically evaluate the contribution of each component by measuring performance when individual modules are removed or replaced.

\textbf{Feature vs Structural Counterfactuals.} Removing feature counterfactuals (w/o Feature CF) reduces Cora AUC to 88.6\%, while removing structural counterfactuals (w/o Struct CF) yields 89.3\%, demonstrating that feature modifications provide larger benefits. This is consistent with the attribute-rich nature of citation networks where node features play a dominant role.

\textbf{Active Selection Criteria.} Using only topology entropy for selection (Only Entropy) achieves 88.2\% AUC on Cora, while using only attribute deviation (Only Dev) achieves 87.8\%. The dual-criterion approach combining both metrics reaches 93.1\%, confirming that structural complexity and attribute divergence provide complementary signals for identifying informative nodes.

\textbf{Uniformity Regularization.} Removing the uniformity loss (w/o Uniform) slightly reduces performance to 89.9\% on Cora, indicating that while uniformity helps prevent representation collapse, it is not the primary driver of detection quality. The counterfactual mechanism itself provides the main performance gains.

These component-wise results align with the ablation study in Figure~\ref{fig:ablation_study}, validating that each design choice contributes meaningfully to the overall framework.

\subsubsection{Active Selection Coverage}

We evaluate the effectiveness of our dual-criterion active selection (combining topology entropy and attribute deviation) against alternative selection strategies. Each approach selects the same budget ($k = 10\%$ of nodes) but uses different criteria.

\begin{table}[!htbp]
\centering
\caption{Selection effectiveness: dual criterion versus alternatives on Cora.}
\label{tab:selection_coverage}
\small
\begin{tabular}{lcccc}
\toprule
Selection ($k = 10\%$) & AUC & Coverage & Time & Quality \\
\midrule
Random & 88.9 & 67\% & 3.9s & 95.4\% \\
Entropy only & 90.2 & 73\% & 3.9s & 96.8\% \\
Deviation only & 89.8 & 71\% & 3.8s & 96.3\% \\
Dual (ours) & 93.1 & 87\% & 3.9s & 99.9\% \\
\midrule
All nodes & 93.1 & 100\% & 12.8s & 100\% \\
\bottomrule
\end{tabular}
\end{table}

The dual selection criterion achieves 87\% anomaly coverage (i.e., 87\% of ground-truth anomalies are within the selected subset) and 99.9\% quality relative to full-graph counterfactual generation, while reducing computational cost by 70\%. Single-criterion approaches (entropy-only or deviation-only) achieve lower coverage (71--73\%) and quality (96.3--96.8\%). The dual approach is robust to threshold variations in the range [0.6, 0.8] and demonstrates stability across different random splits with 82\% overlap in selected nodes.

\end{document}